\documentclass[10pt,twocolumn,letterpaper]{article}

\usepackage{iccv}
\usepackage{times}
\usepackage{epsfig}
\usepackage{graphicx}
\usepackage{amsmath}
\usepackage{amssymb}
\usepackage{microtype}      
\usepackage{multirow}
\usepackage{subfigure}
\usepackage{overpic}
\usepackage{algorithm}
\usepackage{algorithmic}
\usepackage{booktabs}       
\usepackage[misc,geometry]{ifsym}

\usepackage[pagebackref=true,breaklinks=true,letterpaper=true,colorlinks,bookmarks=false]{hyperref}

 \iccvfinalcopy 

\def\iccvPaperID{6352} 

\ificcvfinal\fi

\begin{document}
\title{Reconcile Prediction Consistency for Balanced Object Detection}

\author{Keyang Wang, Lei Zhang$^{(}$\textsuperscript{\Letter}$^)$\\
Learning Intelligence \& Vision Essential (LiVE) Group\\
School of Microelectronics and Communication Engineering, Chongqing University, China\\
{\tt\small \{wangkeyang, leizhang\}@cqu.edu.cn}
}
\maketitle
\ificcvfinal\thispagestyle{empty}\fi

\begin{abstract}
Classification and regression are two pillars of object detectors. In most CNN-based detectors, these two pillars are optimized independently. Without direct interactions between them, the classification loss and the regression loss can not be optimized synchronously toward the optimal direction in the training phase. This clearly leads to lots of inconsistent predictions with high classification score but low localization accuracy or low classification score but high localization accuracy in the inference phase, especially for the objects of irregular shape and occlusion, which severely hurts the detection performance of existing detectors after NMS. To reconcile prediction consistency for balanced object detection, we propose a Harmonic loss to harmonize the optimization of classification branch and localization branch. The Harmonic loss enables these two branches to supervise and promote each other during training, thereby producing consistent predictions with high co-occurrence of top classification and localization in the inference phase. Furthermore, in order to prevent the localization loss from being dominated by outliers during training phase, a Harmonic IoU loss is proposed to harmonize the weight of the localization loss of different IoU-level samples. Comprehensive experiments on benchmarks PASCAL VOC and MS COCO demonstrate the generality and effectiveness of our model for facilitating existing object detectors to state-of-the-art accuracy.
\end{abstract}

\section{Introduction}

Object detection serves as a prerequisite for a broad set of downstream vision applications, such as person re-identification \cite{chang2018multi}, instance segmentation \cite{he2017mask} and  action recognition \cite{gkioxari2018detecting}. With the foundation of deep neural networks, object detection has achieved significant advances in recent years. On the whole, all the detectors that use deep CNN can be divided into two categories: (1) The multi-stage approaches, including \cite{girshick2015fast, ren2015faster, lin2017feature, cai2018cascade,li2019scale,zhou2019ssa}. For the multi-stage object detectors, multi-stage classification and localization are applied sequentially, which make these models more powerful on classification and localization tasks. So these approaches have achieved top performance on benchmark datasets. (2) The one-stage approaches, including \cite{liu2016ssd, redmon2016you, lin2017focal, zhang2018single, wang2020single}. The one-stage approaches apply object classifiers and regressors in a dense manner without object-based pruning. The main advantage of the one-stage detectors is their inference efficiency, but the detection accuracy is usually inferior to the two-stage approaches.

\begin{figure}[t]
\includegraphics[width=0.95\linewidth]{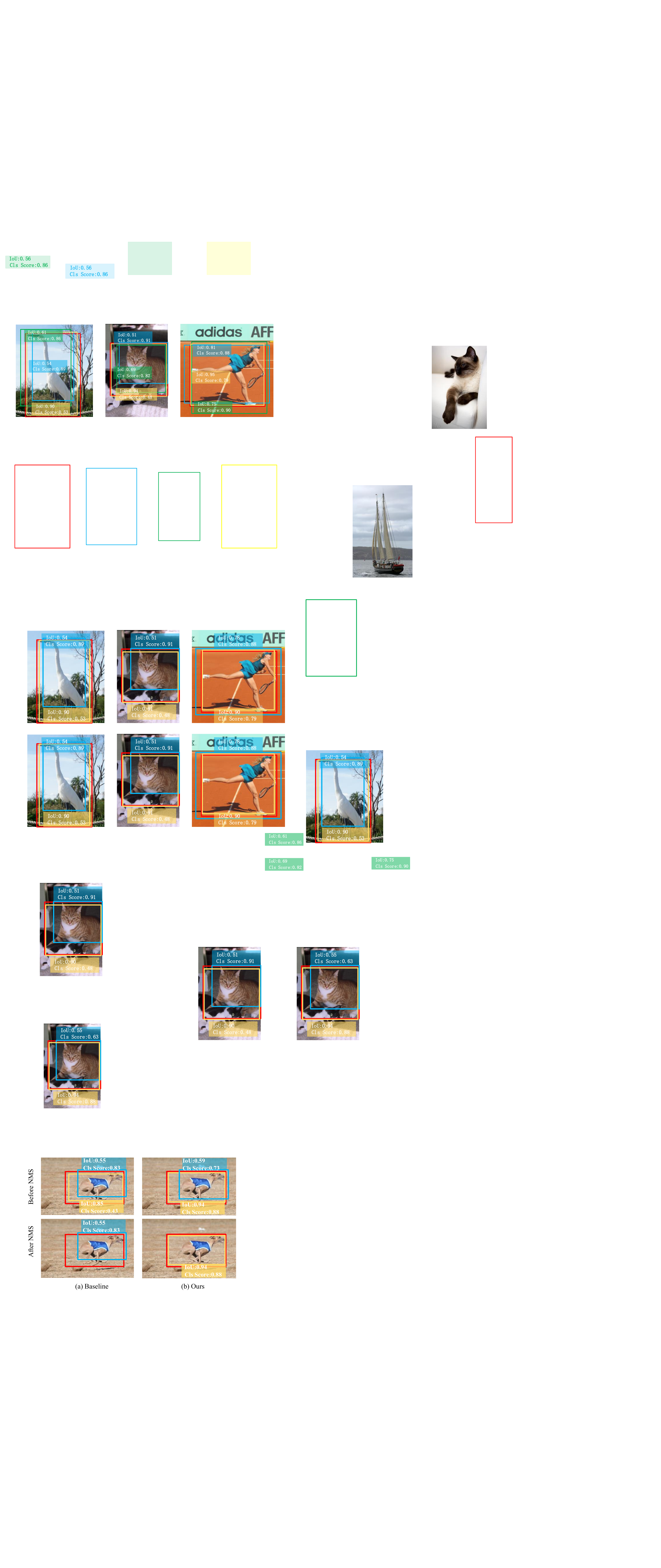}
\vspace{-0.2cm}
\caption{Irregular (e.g., tail) and occluded (e.g., background) object detection outputs of the baseline detector and our detector trained with Harmonic loss. The red bounding boxes denote the ground-truth. The baseline detector (a) produces inconsistent bounding boxes with high classification score but low localization IoU (the blue bbox), or low classification score but high localization IoU (the yellow bbox), which will hurt the detection accuracy after NMS (\emph{i.e.}, suboptimal blue bbox is selected). Our detector (b) produces bounding boxes with high co-occurrence of top classification and localization by harmonizing the optimization of these two branches, leading to better detection results after NMS.
  }
\label{examples}
\vspace{-0.4cm}
\end{figure}

Modern CNN-based object detectors divide the object detection task into two branches, classification branch and regression branch, which are trained with independent objective functions without knowing each other explicitly. Specifically, a classification loss (e.g., cross-entropy loss \cite{girshick2015fast}) is adopted to optimize the classifier and a distance loss (e.g., smoothL1 loss \cite{girshick2015fast}) is used to optimize the regressor. However, since the correlation between classification task and localization task is ignored in existing detectors, the optimization of the classifier and the regressor often can not be synchronous. Specifically, the classification loss drives the model to learn as high classification score as possible for all the positive examples regardless of the localization accuracy, and the regression loss enables the model to output accurate regression offset for each anchor without taking into account the classification score during training. As a result, this leads to serious prediction inconsistency between classification and regression. Specifically, the detector outputs lots of inconsistent detection results that have high classification score but low IoU or low classification score but high IoU in the inference phase. As shown in Fig.~\ref{examples} (a), we show a representative example of inharmonious bounding boxes detected by the baseline. Two inharmonious candidates (a blue bbox and a yellow bbox) are produced by the baseline. Clearly after the NMS procedure, the yellow one with high IoU but low classification score will be suppressed by the less accurate blue one. This means that, for this instance, the detector will end up with a clearly suboptimal detection result (\emph{i.e.}, blue bbox), but not the optimal one (\emph{i.e.}, yellow bbox). Since classification loss and regression loss are independently optimized for each positive sample throughout the training phase, such inconsistency is very common in the testing phase, especially for the objects of irregular shape and occlusion, which will hurt the overall detection performance. However, the prediction inconsistency between tasks is paid less attention in object detection community.

In order to reconcile prediction consistency, we propose a mutually supervised loss, called Harmonic loss, to jointly optimize the classification and regression branches toward the optimal direction. In the training phase, the classification branch will continuously supervise the optimization status of the regression branch through a harmonic factor related to the classification loss, and vice versa. By plugging and playing the Harmonic loss into existing detectors, the network will become training-aware and reduce the training imbalance across tasks by mutual interaction. That is, the easily-trained branch will learn to dynamically facilitate the learning of the other branch towards a progressively harmonic state. As a result, the detectors produce the bounding boxes with high co-occurrence of top classification and localization in the testing, thereby alleviating the prediction inconsistency between these two tasks, as is shown in Fig.~\ref{examples} (b).

Furthermore, during the training phase of the detectors, only positive samples are used to optimize the regression branch, but the IoU distribution of all the positive samples is seriously imbalanced (Please see Fig.\ref{zhu} (a) in the method section). Generally, the number of samples at low IoU levels is significantly larger than that at high IoU levels. This will cause that the localization losses of existing detectors, especially one-stage detectors, are dominated by outliers (low IoU levels) during training phase, which enables the detection model to be biased toward the outliers. In order to alleviate the bias, motivated by focal loss \cite{lin2017focal}, we propose a simple but effective Harmonic IoU (HIoU) loss to dynamically harmonize the contribution of each kind of examples in training phase. We up-weight the weights of localization loss of examples with high IoU while suppressing the weights of examples with low IoU by a dynamic scaling factor. By embedding this HIoU loss into our proposed Harmonic loss, the contribution of each level of samples is balanced and the training is more effective and stable.

By embedding the HIoU loss into our proposed Harmonic loss, a Harmonic detection (HarmonicDet) loss is formulated. We plug and play our HarmonicDet loss into both two-stage and single-stage detection frameworks on the PASCAL VOC \cite{everingham2007pascal} and MS COCO \cite{lin2014microsoft} benchmarks. The detection results demonstrate the effectiveness and generality of our approach in improving existing detectors. In summary, this paper makes three contributions:

\hangafter=1
\setlength{\hangindent}{2.0em}
1. We propose a Harmonic loss to jointly optimize the classification and regression branch toward the optimal direction in the training phase, thereby reconciling prediction consistency
between classification and localization in the testing phase.

\hangafter=1
\setlength{\hangindent}{2.4em}
2. We introduce a Harmonic IoU (HIoU) loss to harmonize the weight of the localization loss of different IoU-level samples, which prevents the localization loss from being dominated by outliers and ensures accurate bounding box regression ability of the detector.

\hangafter=1
\setlength{\hangindent}{2.4em}
3. The proposed losses can be easily plugged and played into different state-of-the-art detection algorithms, achieving significant performance gains on the PASCAL VOC dataset and MS COCO dataset.

\section{Related Work}
\vspace{-0.2cm}
\textbf{Architecture design for object detection}. Benefit from the Deep ConvNets, CNN has achieved a great success in the object detection field. All the CNN-based detectors can be roughly divided into two categories, \emph{i.e.,} two-stage detectors and one-stage detectors. The two-stage detectors consist of two parts. The first part is responsible for generating a set of candidate object proposals, e.g., Selective Search \cite{uijlings2013selective}, EdgeBoxes \cite{zitnick2014edge}. The second part determines the accurate object regions and the corresponding class labels according to the candidate object proposals. Its descendants (e.g., Fast R-CNN \cite{girshick2015fast}, Faster R-CNN \cite{ren2015faster}, R-FCN \cite{dai2016r}, FPN \cite{lin2017feature}, Mask RCNN \cite{he2017mask}) achieve dominant performance on several challenging datasets. In contrast, one-stage approaches like SSD \cite{liu2016ssd} directly predict the class scores and location offsets of the default bounding boxes, without the region proposal step. After that, more advanced one-stage detectors (e.g., RetinaNet \cite{lin2017focal}, RefineDet \cite{zhang2018single}, RFB \cite{liu2018receptive}) are proposed to achieve high detection accuracy. In order to solve the problem that the detection result is too sensitive to the size of the anchor and simultaneously avoid the complex IoU computation between anchor boxes and ground-truth boxes during training, some anchor-free detectors are proposed, including CornerNet \cite{law2018cornernet}, FCOS \cite{tian2019fcos}, CenterNet \cite{duan2019centernet}, FASF \cite{Zhu_2019_CVPR}.

\textbf{Loss design for accurate object detection}. In order to strengthen the correlation between classification and localization tasks and get accurate bounding boxes, some detectors have also made some representative explorations. IoU-Net \cite{Jiang_2018_ECCV} integrates the IoU predictor into existing object detectors to predict the IoU between each detected bounding box and the matched ground-truth. Simultaneously, an IoU-guided NMS is proposed to resolve the misalignment between classification confidence and localization accuracy.
PISA \cite{Cao_2020_CVPR} uses the classification probabilities to reweight the contribution of each example to the regression loss, such that the regression branch can be strengthened for improving detection performance.
In order to balance the contribution of each kind of samples to the localization branch, some sample-adaptive localization losses are studied. GHM \cite{li2019gradient} analyzes the sample imbalance in one-stage detectors in term of gradient norm distribution and propose a gradient aware localization loss GHM-R. Libra R-CNN \cite{pang2019libra} claims that the small gradients produced by the easy samples may be neglected comparing to the large gradients produced by the outliers and a balanced $\mathcal{L}_1$ loss is proposed to increase the gradient of easy samples and keep the outliers unchanged.

\section{Proposed Approach}
\vspace{-0.2cm}
In this section, we will firstly introduce the Harmonic loss in section~\ref{CD_loss}. In particular, the in-depth insight on how it can harmonize the optimization from the loss and its gradient is presented. Then we further elaborate the proposed Harmonic IoU (HIoU) loss in section~\ref{wiou_loss}.

\subsection{Harmonic Loss}
\vspace{-0.2cm}
\label{CD_loss}
We firstly revisit the standard loss of object detector, as shown in Eq.(\ref{eq1}).
\begin{small}
\begin{equation}
\begin{split}
\label{eq1}
\mathcal{L}_{Det}=&\frac{1}{N}(\sum\limits_{i \in Pos}^N (CE(p_i,y_i)+\mathcal{L}(d_i,\hat{d_i}))+\sum\limits_{j \in Neg}^M CE(p_j,y_j))
\end{split}
\vspace{-0.8cm}
\end{equation}
\end{small}
where \emph{N} and \emph{M} is the number of \emph{Positive} and \emph{Negative} samples, resp. And $p_i$ and $y_i$ denotes the predicted classification score and the corresponding ground truth class, resp. $d_i$ and $\hat{d_i}$ denotes the output regression offset and target offset, resp. CE($\cdot$) is the cross entropy loss and $\mathcal{L}(\cdot)$ is the commonly used Smooth L1 loss,

We can clearly find that for each positive sample, the classification and localization branches are trained with independent objective functions without knowing each other explicitly, which causes serious prediction inconsistency between classification and regression tasks. For example, for classification branch, the CE loss drives the model to learn as high classification score as possible for all samples regardless of the localization accuracy during training. As a result, the predicted classification scores are independent of the localization accuracy and there must be many inconsistent detections that have high classification score but low IoU or low classification score but high IoU, which deteriorates the overall detection performance after NMS.

In order to avoid these inconsistent detections, we propose a Harmonic loss to harmonize the training of classification and regression branches and strengthen the correlation between classification score and localization accuracy. Since only positive samples need to be regressed, we only apply Harmonic loss on positive samples. So given a positive sample $x_i$, the Harmonic loss can be defined as follows:
\vspace{-0.1cm}
\begin{equation}
\begin{split}
\label{eq2}
\mathcal{L}_{Har}^i=&(1+\beta_r)CE(p_i,y_i)+(1+\beta_c)\mathcal{L}(d_i,\hat{d_i}) \\
\end{split}
\end{equation}
where $\beta_r$, $\beta_c$ are two key but dynamic harmonic factors in our method, which can be defined as follows:
\vspace{-0.1cm}
\begin{equation}
\begin{split}
\label{eq_cor}
&\beta_r=e^{-\mathcal{L}(d_i,\hat{d_i})} \hspace{0.3cm}, \hspace{0.3cm} \beta_c=e^{-CE(p_i,y_i)}\\
\end{split}
\vspace{-0.3cm}
\end{equation}

From Eq.(\ref{eq2}) and Eq.(\ref{eq_cor}), we can clearly find that there are two mutually interacted parts in our Harmonic loss. We use a mutual supervision mechanism to harmonize the optimization of classification and regression branch. For classification task, a regression loss derived harmonic factor $(1+\beta_r)$ is assigned to dynamically supervise the optimization of the classification branch. In other words, a regression-aware classification loss is devised to optimize the classification branch. Similarly, for regression task, a dynamic factor $(1+\beta_c)$ related to the classification loss is used to supervise the optimization of regression loss, which means that the regression loss is classification-aware. In the following, in-depth insight and detailed analysis on why the Harmonic loss can harmonize the classification and regression branch are elaborated from the loss function and the gradient. And due to space limitation, we give the convergence analysis of our harmonic loss in the \textbf{supplementary material}.

\textbf{Effectiveness analysis from the loss function}. Given an object $x_i$, and suppose the predicted class score w.r.t. its ground truth class to be $p_i$ and the location accuracy w.r.t. the ground truth bounding box to be $IoU_i$. We may face four different situations during the optimization process:
\textbf{(1)} \emph{classification is good} ($p_i\uparrow$) \emph{and regression is good} ($IoU_i\uparrow$). This means that the sample $x_i$ is the prime sample for learning high classification scores and accurate localization. So we simultaneously assign a large harmonic factor $\beta_r$ for classification loss according to the state of regression, and a large harmonic factor $\beta_c$ for regression loss according to the state of classification.
\textbf{(2)} \emph{classification is good} ($p_i\uparrow$) \emph{but regression is bad} ($IoU_i\downarrow$). This is one of the inharmonious training situations that we need to solve. Since the regression task for this positive sample is poor, we assign a large harmonic factor $\beta_c$ for regression loss according to the the state of classification and a small harmonic weight $\beta_r$ for  classification loss according to the state of regression. Intuitively, the network will reinforce attention to primarily optimize the regression branch.
In addition, since the dynamic harmonic factor $\beta_c$ is derived by the classification loss, this means that the regressor can always perceive the classification status. In other words, the classification loss will also supervise the optimization of the regression branch. This is also true for classification branch. As a result, these two branches will supervise and promote each other, and optimize simultaneously toward the optimal direction. The disharmony between classification and regression is therefore alleviated.
\textbf{(3)} \emph{classification is bad} ($p_i\downarrow$) \emph{but regression is good} ($IoU_i\uparrow$). This is another inharmonious training situation. Contrary to situation (2), a small harmonic weighting factor $\beta_c$ and a large harmonic weighting factor $\beta_r$ is assigned for regression loss and classification loss, respectively. As a result, the network primarily optimize the classification branch and alleviate the disharmony between the two tasks. From situation (2) and (3), we see that in the Harmonic loss the easily-trained branch is always enabled to dynamically facilitate the learning of the other branch.
\textbf{(4)} \emph{classification is bad} ($p_i\downarrow$) \emph{and regression is bad} ($IoU_i\downarrow$). This means that the positive sample $x_i$ is a hard positive sample. We claim that those hard positive samples are not the prime samples and have little effect on the optimization of the detectors. So we simultaneously decrease the weights $\beta_r$, $\beta_c$ of the classification loss and localization loss, resp. By applying the Harmonic loss to optimize the existing detectors during training, these detectors can naturally harmonize the classification score and localization accuracy in the testing phase.

\textbf{Effectiveness analysis on classification task from the gradient}. We have described in detail how to apply the Harmonic loss to achieve the interaction and harmony between classification task and regression task from the aspect of loss function. In the following, we will further discuss how our Harmonic loss harmonizes the training of the two tasks from the aspect of gradient. For classification task, we calculate the partial derivative of Eq.(\ref{eq2}) with respect to the predicted score $p_i$ of the corresponding ground truth class as:
\begin{equation}
\begin{split}
\vspace{-0.3cm}
\label{eq3}
\frac{\partial \mathcal{L}_{Har}^i}{\partial p_i}=(1+\beta_r)\frac{\partial CE(p_i,y_i)}{\partial p_i}+\mathcal{L}(d_i,\hat{d_i})\frac{\partial \beta_c}{\partial p_i}\\
\end{split}
\end{equation}
where CE($\cdot$) is the cross-entropy loss and $y_i$ is the one-hot label, and there is
\begin{equation}
\begin{split}
\label{eq4}
&CE(p_i,y_i)=-y_i\log(p_i)=-\log(p_i),\\
 &\beta_c=e^{-CE(p_i,y_i)}=e^{y_i\log(p_i)}=p_i
\end{split}
\end{equation}
By substituting Eq.(\ref{eq4}) into Eq.(\ref{eq3}), we can have
\begin{equation}
\begin{split}
\label{eq5}
\frac{\partial \mathcal{L}_{Har}^i}{\partial p_i}=-(1+\beta_r)\frac{1}{ p_i}+\mathcal{L}(d_i,\hat{d_i})\\
\end{split}
\end{equation}
By substituting $\beta_r=e^{-\mathcal{L}(d_i,\hat{d_i})}$ into Eq.(\ref{eq5}), we can have
\begin{equation}
\begin{split}
\label{eq6}
\frac{\partial \mathcal{L}_{Har}^i}{\partial p_i}=\mathcal{L}(d_i,\hat{d_i})-\frac{(1+e^{-\mathcal{L}(d_i,\hat{d_i})})}{ p_i}
\end{split}
\end{equation}

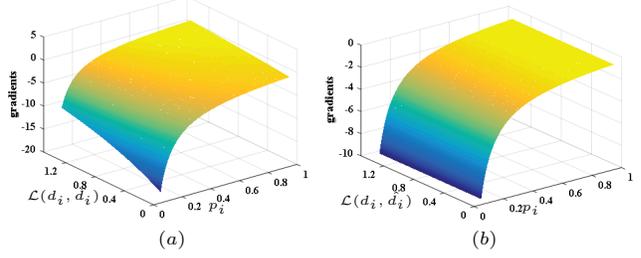
\begin{figure}
 \centering
 \begin{overpic}[width=1.0\linewidth]{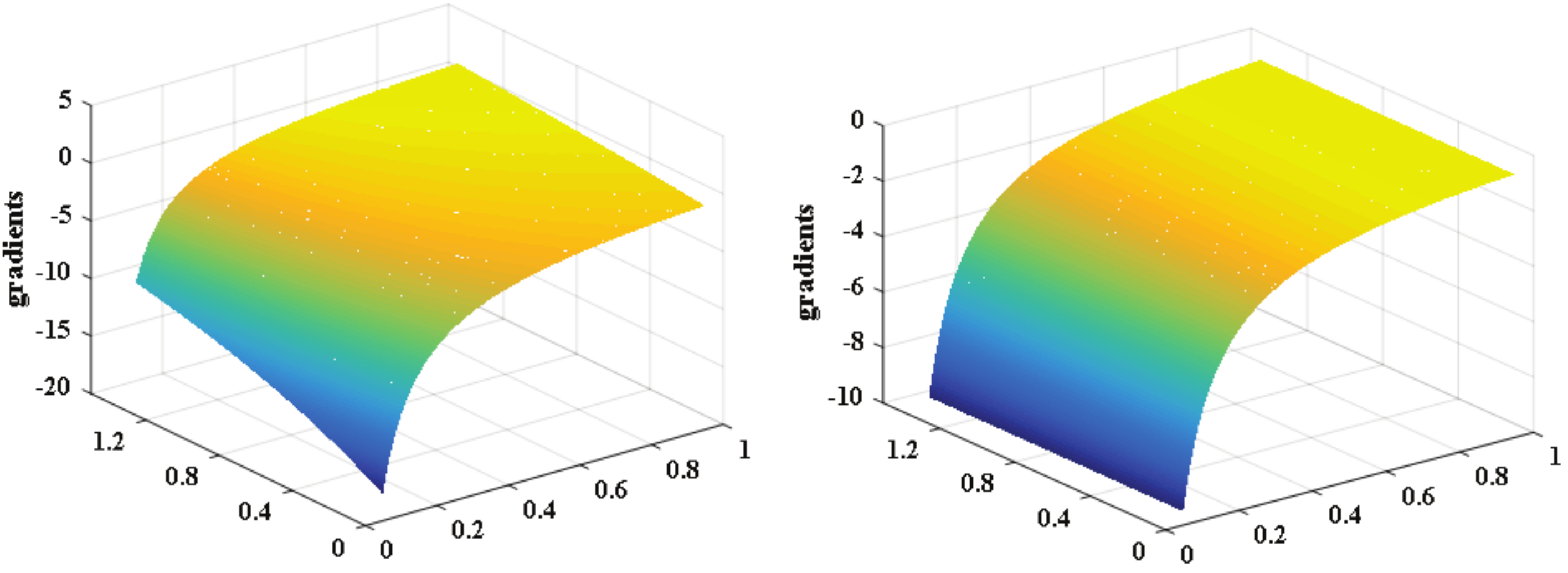}
 \put(3.1,3.7){\tiny$\mathcal{L}{(d_i,\hat{d_i})}$}
 \put(32,2.3){\tiny $p_{i}$}
 \put(24,-3.4){\scriptsize $(a)$}
 \put(53.0,3.0){\tiny  $\mathcal{L}{(d_i,\hat{d_i})}$}
 \put(82,2.1){\tiny $p_{i}$}
 \put(74,-3.4){\scriptsize $(b)$}
 \end{overpic}
 \vspace{0cm}
 \caption{Visualization of the gradients of detection losses with respect to the predicted score of the corresponding ground truth class $p_i$. (a) is the gradient of our Harmonic loss with respect to $p_i$, $\frac{\partial \mathcal{L}_{Har}^i}{\partial
  p_i}$. (b) is the gradient of the standard detection loss (cross-entropy loss plus smooth L1 loss) with respect to $p_i$.}
  \label{grad}
  \vspace{-0.3cm}
\end{figure}

We can clearly find that the partial derivative of our Harmonic loss to the classification probability $p_i$ is correlated to the regression loss, which means that the classification branch will be supervised by the regression loss during the optimization process. In order to further illustrate the effect of the localization loss to the gradient in Eq.(\ref{eq6}), we visualize the gradients of the standard detection loss and the proposed Harmonic loss with respect to $p_i$, respectively, in Fig.~\ref{grad}. For the standard detection loss, the gradient with respect to $p_i$ does not change with different location loss values as Fig.~\ref{grad} (b) shows, which means the optimization of classification is absolutely independent of the regression task. But for our Harmonic loss, the gradient is a function simultaneously determined by the two variables $p_i$ and $\mathcal{L}(d_i,\hat{d_i})$. In other words, for each positive sample, the localization loss $\mathcal{L}(d_i,\hat{d_i})$ will supervise the optimization of the classification branch during training phase of the Harmonic loss. Specifically, in Eq.(\ref{eq6}), there is a proportional correlation between $\mathcal{L}(d_i,\hat{d_i})$ and the loss gradient w.r.t. $p_i$. But since this gradient value is always negative during the training process, the absolute value of this gradient decreases as the regression loss increases, as is shown in Fig.~\ref{grad} (a). This means that the gradient will suppress the classification scores of samples with low regression quality.

\textbf{Effectiveness analysis on regression task from the gradient}. The mechanism of our Harmonic loss for regression task is similar to classification task. For regression task,the gradient of Harmonic loss with respect to the predicted regression offset $(d_i-\hat{d_i})$ is also supervised by the classification score $p_i$. Due to space limitation, we give this derivation process in the \textbf{supplementary material}.

\textbf{Task-Contrastive Loss}. In order to further reduce the gap between the two tasks, we introduce a simple Task-Contrastive (TC) loss, as defined in Eq.(\ref{eq10}), to our Harmonic loss. We directly reduce the contrastive distance between the predicted score $p_i$ and the localization accuracy $IoU_i$ for the sample $x_i$ w.r.t. its ground truth class and bounding box. A margin is adopted because we think that the gap between classification and regression can be ignored when the distance between $p_i$ and $IoU_i$ is smaller than the margin. Additionally, we assign an information entropy-guided weight factor $\frac{1}{1+\beta_e}$ for each sample. Specifically, when the entropy is large, it means that the classification uncertainty is large, so a small weight, $\frac{1}{1+\beta_e}$, is assigned, and vice versa.
\vspace{-0.2cm}
\begin{equation}
\begin{split}
\label{eq10}
&\mathcal{L}_{TC}^i=\frac{1}{1+\beta_e}\max(0,|p_i-IoU_i|-margin)\\
&\beta_e=e^{-\sum\limits_{k=1}^C p_k\log(p_k)}
\end{split}
\vspace{-0.5cm}
\end{equation}
where \emph{C} is the number of classes, and the \emph{margin} is set as 0.2 in our experiments.

\subsection{Harmonic IoU Loss}
\label{wiou_loss}
The IoU based localization loss \cite{yu2016unitbox} has proved to be a suitable choice to obtain the optimal IoU metric, which is written as:
\begin{small}
\begin{equation}
\vspace{-0.2cm}
\label{eq7}
\mathcal{L}_{IoU}^{i}=1-IoU_{i}
\end{equation}
\vspace{-0.3cm}
\end{small}

\begin{figure}
 \centering
 \begin{overpic}[width=1.0\linewidth]{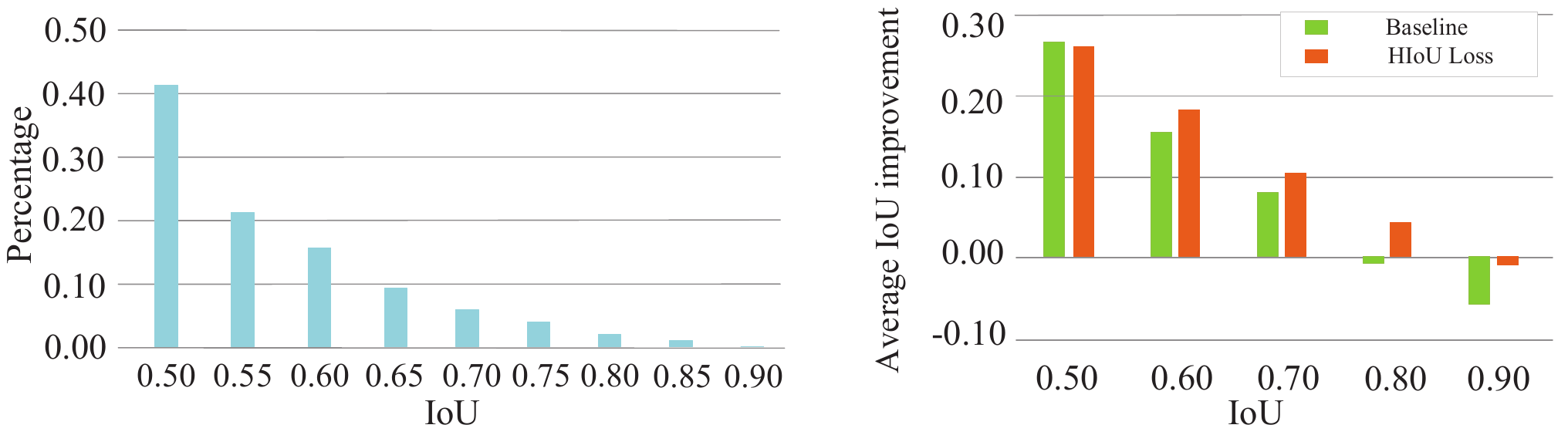}
 \vspace{-0.1cm}
 \put(26,-3.4){\scriptsize (a)}
 \put(76,-3.4){ \scriptsize(b)}
 \end{overpic}
 \vspace{0.0cm}
 \caption{Two histograms of the positive samples. (a) is the IoU distribution of the 100k positive samples before refinement. (b) is the average localization improvement of proposals from different IoU intervals after refinement.}
 \label{zhu}
 \vspace{-0.2cm}
\end{figure}

However, the typical IoU loss treats all the positive samples equally and neglects a fact that IoU distribution is seriously imbalanced. As shown in Fig.~\ref{zhu} (a), we visualize the IoU distribution of the 100k positive samples. We can find that the number of samples at low IoU levels is much larger than that at high IoU levels. Needless to say, the hard samples at low IoUs dominate the gradients of the localization loss in the training phase, which enables the detection model to be biased toward the hard samples (low IoU levels). As shown in Fig.~\ref{zhu} (b), we visualize the average localization improvement of proposals from different IoU intervals after refinement. We can find that, with the increase of IoU, the gain of refinement gets smaller and the performance even gets worse at high IoU levels (IoU=0.8, 0.9). This is because the regressor always outputs large offset in the testing phase as it is biased towards hard samples. In order to alleviate the bias, motivated by focal loss \cite{lin2017focal}, we introduce a Harmonic IoU (HIoU) loss, defined as:
\vspace{-0.2cm}
 \begin{small}
\begin{equation}
\vspace{-0.2cm}
\begin{split}
\label{eq8}
 \mathcal{L}_{HIoU}^i=(1+IoU_i)^\gamma(1-IoU_i)
\end{split}
\end{equation}
\end{small}
$\gamma$ is the tunable focusing parameter. A dynamic factor $(1+IoU_i)^\gamma$ that can automatically reflect the localization accuracy is used to harmonize the weight of the localization loss of different IoU-level samples. Specifically, we use the factor to increase the weights of localization loss of examples with high IoU while suppressing the weights of examples with low IoU. In the end, the contribution of each kind of examples is balanced and the bias of regression can be effectively alleviated. As shown in Fig.~\ref{zhu} (b), the model trained with our Harmonic IoU loss can prevent the localization loss from being dominated by outliers during training phase and ensure accurate bounding box regression ability of the whole detector. Let us take the samples with IoU greater than 0.8 before refinement as an example. The original localization loss causes negative regression, but our Harmonic IoU loss can effectively avoid this situation. It should be noted that in order to ensure that the Harmonic IoU loss is always monotonic, the value of $\gamma$ needs to satisfy $\gamma \le 1$, we give the derivation in the \textbf{supplementary materials}.

\textbf{Full Localization Loss.} It should be noted that the SmoothL1 loss can directly optimize the distance between central points of the predicted boxes and GT boxes, thereby accelerating the convergence speed of training. So we do not replace the SmoothL1 loss in our Harmonic loss with HIoU loss, but add the HIoU loss as a part of the localization loss. So the full localization loss can be defined as follows:
\vspace{0.2cm}
\begin{equation}
\begin{split}
\label{eq_loc}
\mathcal{L}_{loc}^{i}=\mathcal{L}(d_i,\hat{d_i})+\alpha \mathcal{L}_{HIoU}^i \\
\end{split}
\vspace{0.2cm}
\end{equation}
where $\alpha$ is a trade-off parameter.

\subsection{Overall Detection Loss}
By replacing localization loss $\mathcal{L}(d_i,\hat{d_i})$ in Eq.~(\ref{eq2}) with the full localization loss in Eq.~(\ref{eq_loc}) and embedding the TC loss as defined in Eq.~(\ref{eq10}), the overall \textbf{Harmonic} \textbf{Det}ection loss for positive sample $x_i$ can be formulated as follows:
\begin{small}
\begin{equation}
\begin{split}
\label{eq9}
\mathcal{L}_{HarDet}^i=&(1+\beta_r)CE(p_i,y_i)+(1+\beta_c)\mathcal{L}_{loc}^{i}+\mathcal{L}_{TC}^i\\
\end{split}
\end{equation}
\end{small}
where $\beta_r=e^{-\mathcal{L}_{loc}^{i}}, \beta_c=e^{-CE(p_i,y_i)}$ are the two key dynamic harmonic factor derived in Eq.(\ref{eq_cor}). By further adding the classification loss of negative samples, we can get the whole objective function as follows:
\begin{small}
\begin{equation}
\begin{split}
\label{eq11}
\mathcal{L}_{HarDet}=\frac{1}{N}(\sum\limits_{i \in Pos}^N \mathcal{L}_{HarDet}^i+ \sum\limits_{j \in Neg}^M CE(p_j,y_j))
\end{split}
\end{equation}
\end{small}
\begin{table*}

  \caption{Main results on PASCAL VOC 2007 \texttt{test} set. Four different detectors are adopted in our experiments.
  }
  \footnotesize
  \label{sample-table1}
  \centering
  \begin{tabular}{c|c|c|p{0.8cm}p{0.8cm}p{0.8cm}p{0.8cm}p{0.8cm}}
    \toprule
    Method      &     Backbone         &   AP& AP$_{50}$ & AP$_{60}$ & AP$_{70}$ & AP$_{80}$ & AP$_{90}$   \\
    \hline


    \footnotesize\emph{two-stage detectors:}& & &&&&&\\
    Faster R-CNN              & ResNet-50-FPN   &51.7&80.9&76.0&65.1&43.8&11.1  \\
    Faster R-CNN             & ResNet-101-FPN   &54.1&81.8&77.4&67.8&48.5&15.0  \\
    \hline
    Faster R-CNN w/HarmonicDet Loss           & ResNet-50-FPN   &\textbf{54.2}(+2.5)&81.7&77.5&67.3&48.2&14.7  \\
    Faster R-CNN w/HarmonicDet Loss        & ResNet-101-FPN   &\textbf{56.5}(+2.4)&82.6&78.7&70.1&52.5&18.2 \\
    \hline
    \hline
    \footnotesize\emph{one-stage detectors:}& & &&&\\
    SSD300             & ResNet-50-FPN  &51.0&78.9&73.4&63.3&44.1&14.6   \\
    SSD512             & ResNet-50-FPN  &53.3&81.5&76.8 &66.6&47.3&15.3 \\
    DSSD320              & ResNet-50    &52.3&79.8&75.0 &64.5&46.2&16.2 \\
    DSSD512              & ResNet-50    &54.2&81.8&77.7 &67.4&50.0&15.7 \\
    RefineDet320  & ResNet-50           &53.4&80.2&76.0 &65.5&48.0& 17.5\\
    RefineDet512  & ResNet-50           &56.3&82.4&79.3 &69.8&52.7& 19.6\\
    \hline
    SSD300  w/HarmonicDet Loss          & ResNet-50-FPN  &\textbf{54.5} (+3.5)&79.7&74.9 &65.8&50.2&21.2  \\
    SSD512  w/HarmonicDet Loss           & ResNet-50-FPN  &\textbf{56.7} (+3.4)&81.8&78.0 &68.9&52.7&22.0  \\
    DSSD320 w/HarmonicDet Loss             & ResNet-50     &\textbf{55.7} (+3.4)&80.0&75.7 &66.9&52.1&23.7 \\
    DSSD512 w/HarmonicDet Loss            & ResNet-50     &\textbf{57.3} (+3.1)&82.4&78.6 &69.0&53.4&23.1 \\
    RefineDet320 w/HarmonicDet Loss  & ResNet-50          &\textbf{56.9} (+3.5)&80.2&76.2 &67.5&53.2& 27.2\\
    RefineDet512 w/HarmonicDet Loss   & ResNet-50           &\textbf{60.4} (+4.1)&83.0&79.7 &72.0&58.0& 29.7\\
    \bottomrule
  \end{tabular}
  \vspace{-0.4cm}
\end{table*}
\vspace{-0.7cm}
\section{Experiments}
\vspace{-0.2cm}
\subsection{Experimental Setting}
\vspace{-0.2cm}
\textbf{Datasets and Baselines}. We integrate the proposed loss function into six popular one-stage and two-stage detectors including SSD, DSSD, RefineDet, RetinaNet, Faster R-CNN and Mask R-CNN, and conduct experiments on two benchmarks,\emph{ i.e.}, PASCAL VOC and MS COCO.

\textbf{Backbones}. ResNet-50 \cite{he2016deep}, ResNet-101 \cite{he2016deep}, ResNeXt-101 \cite{Xie_2017_CVPR}, pretrained on the standard classification task \cite{russakovsky2015imagenet}, are used as our backbone networks. Specifically, ResNet-50 and ResNet-101 are mainly used in the experiments of PASCAL VOC dataset. ResNet-50, ResNet-101 and ResNeXt-101 are all used in the experiments of MS COCO dataset.
\vspace{-0.3cm}
\subsection{Experiments on PASCAL VOC}
\vspace{-0.1cm}
For PASCAL VOC dataset, three one-stage detectors (SSD, DSSD, RefineDet) and one two-stage detector (Faster R-CNN) are adopted as the detectors in our experiments. All models are trained on the union of the VOC 2007 \texttt{trainval} and VOC 2012 \texttt{trainval} datasets, and tested on the VOC 2007 \texttt{test} set. In order to prove that our model can regress more accurate bounding boxes, a stricter COCO-style Average Precision (averaged AP at IoUs from 0.5 to 0.9) metric is adopted on the PASCAL VOC dataset.

From Table~\ref{sample-table1}, we can see that our HarmonicDet loss achieves consistently significant AP improvements on both the one-stage and two-stage detectors, indicating its effectiveness and generality. For two-stage detectors, our HarmonicDet loss can improve the AP by 2.5$\%$ compared with the baselines Faster R-CNN with ResNet-50-FPN backbone. Even with a deeper backbone like ResNet-101-FPN, similar improvements are observed. For one-stage detectors, our HarmonicDet loss improves SSD, DSSD and RefineDet by 3.5$\%$, 3.4$\%$ and 3.5$\%$, respectively, when we adopt the small input size. Even with a larger input size (512$\times$512), our HarmonicDet loss can also improve SSD, DSSD and RefineDet by 3.4$\%$, 3.1$\%$ and 4.1$\%$, respectively. Especially, the improvement for AP at higher IoU threshold (0.8, 0.9) is very significant. When the IoU threshold is 0.8, our HarmonicDet loss can improve the AP by 5$\%$ $\sim$ 6$\%$. When the IoU threshold is 0.9, our HarmonicDet loss can even improve the AP by 10$\%$. This proves that the detectors trained with our HarmonicDet loss can regress more accurate bboxes.
\subsection{Experiments on MS COCO}
\vspace{-0.1cm}
For MS COCO dataset, three one-stage detectors (SSD, RefineDet, RetinaNet) and two two-stage detector(Faster R-CNN, Mask R-CNN) are adopted as the detectors in our experiments. The results on MS COCO dataset are shown in Table~\ref{sample-table2}, our HarmonicDet loss also achieves consistently significant AP improvements on both the one-stage and two-stage detectors. Specifically, on SSD300 and SSD512, the gain is 1.9$\%$ and 1.7$\%$, respectively, with the ResNet-50-FPN backbone. On RefineDet320 and RefineDet512, the gain is 1.4$\%$ and 2.4$\%$, respectively, with the ResNet-101 backbone. For the RetinaNet, our approach can improve the AP by 1.5$\%$ with ResNet-50-FPN backbone. Even with a strong backbone like ResNeXt-101-32x8d, our model also outperforms the baseline by 1.4$\%$ AP. For two-stage detector, HarmonicDet loss can improve the AP of Faster R-CNN, Mask R-CNN by 1.8$\%$, 1.6$\%$, respectively, with ResNet-50-FPN backbone. When we adopt the ResNeXt-101-FPN as the backbone, our method can achieve 46.9$\%$ AP, which is better than some state-of-the-art detectors, such as Cascade R-CNN (42.8$\%$), FCOS (42.7$\%$) and FASF (42.9$\%$).
\vspace{-0.1cm}

\begin{table*}
  \caption{Main Results of our experiments on MS COCO \texttt{test-dev} set. The $\ddagger$ means multi-scale testing}
  \footnotesize
  \label{sample-table2}
  \centering
  \begin{tabular}{c|c|c|cc|ccc}
    \toprule
    Method           &   Backbone      & AP & AP$_{50}$ & AP$_{75}$ &  AP$_{S}$& AP$_{M}$ &AP$_{L}$ \\
    \hline
    \footnotesize\emph{state-of-the-art detectors:}&  &&&&&&\\
    FCOS \cite{tian2019fcos}&ResNeXt-101-FPN &42.7 &62.2 &46.1 &26.0 &45.6 &52.6 \\
    FASF  \cite{Zhu_2019_CVPR} &ResNeXt-101-FPN     &42.9 &63.8 &46.3 &26.6 &46.2 &52.7  \\
    Cascade R-CNN\cite{cai2018cascade}  &ResNet-101-FPN &42.8 &62.1 &46.3 &23.7 &45.5 &55.2                    \\
    \hline
    \hline
    \footnotesize\emph{one-stage detectors:}&  &&&&&&\\
    SSD300             & ResNet-50-FPN    &29.5 & 49.0&30.5& 11.0 & 32.6 & 47.0    \\
    SSD512           & ResNet-50-FPN      & 34.4 &55.0& 36.5&16.3 & 39.6 & 50.1   \\
    RefineDet320   & ResNet-101& 32.0 &51.4& 34.2 &10.5& 34.7 &50.4                  \\
    RefineDet512  &ResNet-101 &36.4 &57.5 &39.5 &16.6 &39.9 &51.4                    \\
    RetinaNet            & ResNet-50-FPN      &36.1 &56.0 &38.3 &19.8 &38.9 &45.0  \\
    RetinaNet             & ResNeXt-101-FPN      &39.8 &59.9 &42.9 &22.5 &43.1 &50.9  \\
    \hline
    SSD300 w/HarmonicDet Loss            & ResNet-50-FPN    &\textbf{31.4}(+1.9) & 49.3&32.9& 11.1 & 33.3 & 49.3    \\
    SSD512 w/HarmonicDet Loss          & ResNet-50-FPN      & \textbf{36.1}(+1.7) &55.2& 38.3&16.8 & 40.6 & 51.2   \\
    RefineDet320 w/HarmonicDet Loss   & ResNet-101& \textbf{33.4}(+1.4) &51.6& 35.4 &11.1& 35.2 &52.3                  \\
    RefineDet512 w/HarmonicDet Loss  &ResNet-101 &\textbf{38.8} (+2.4)&56.5 &41.8 &17.8 &44.4 &54.8                    \\
    RetinaNet  w/HarmonicDet Loss            & ResNet-50-FPN      &\textbf{37.6}(+1.5) &56.0 &40.1 &20.0 &39.6 &48.7  \\
    RetinaNet  w/HarmonicDet Loss            & ResNeXt-101-FPN      &\textbf{41.2}(+1.4) &60.3 &44.2 &23.2 &43.8 &52.4  \\
    \hline
    \hline
    \footnotesize\emph{two-stage detectors:}&  &&&&&&\\
     Faster R-CNN             & ResNet-50-FPN  &37.5 & 59.0& 41.0&21.4&40.1&49.5   \\
     Faster R-CNN             & ResNeXt-101-FPN &41.4 & 63.6& 45.3&25.4&45.5& 52.3   \\
     Mask R-CNN             & ResNet-50-FPN  &38.5 & 59.5& 41.9&21.9&40.8&50.1   \\
     Mask R-CNN             & ResNeXt-101-FPN &42.3 & 64.1& 46.2&26.1&45.9& 54.3   \\
    \hline
     Faster R-CNN  w/HarmonicDet Loss           & ResNet-50-FPN  &\textbf{39.3}(+1.8) & 59.8&42.7&22.0&41.4& 50.0   \\
     Faster R-CNN w/HarmonicDet Loss            & ResNeXt-101-FPN  &\textbf{43.0}(+1.6) & 64.0& 47.0&26.0&46.1& 53.6  \\
     Mask R-CNN  w/HarmonicDet Loss            & ResNet-50-FPN  &\textbf{40.1}(+1.6) & 60.5&43.7&22.8&42.3& 50.7   \\
     Mask R-CNN w/HarmonicDet Loss            & ResNeXt-101-FPN  &\textbf{44.0}(+1.7) & 64.9& 48.1&27.0&47.0& 55.3  \\
     Mask R-CNN w/HarmonicDet Loss $\ddagger$ & ResNeXt-101-FPN  &\textbf{46.9} & 66.0& 50.8&30.2&50.2& 57.9  \\
    \bottomrule
  \end{tabular}
  \vspace{-0.4cm}
\end{table*}
\subsection{Analysis of Prediction Consistency}
\begin{figure}[h]
  \vspace{-0.3cm}
  \centering
  \subfigure[]{
  \begin{minipage}{0.47\linewidth}
  \centering
  \includegraphics[width=1.59in]{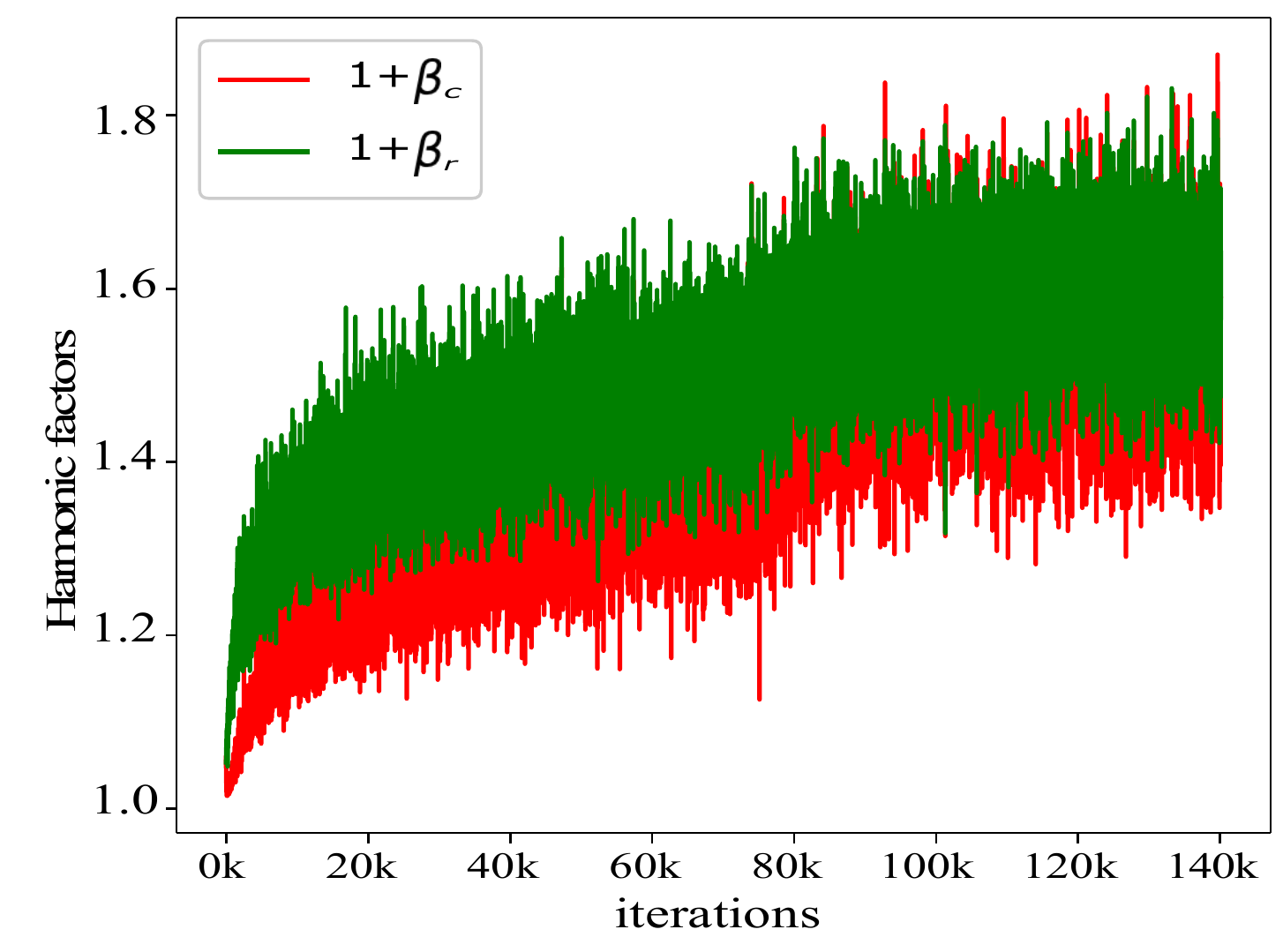}
  \end{minipage}
  }
  \subfigure[]{
  \begin{minipage}{0.47\linewidth}
  \centering
  \includegraphics[width=1.59in]{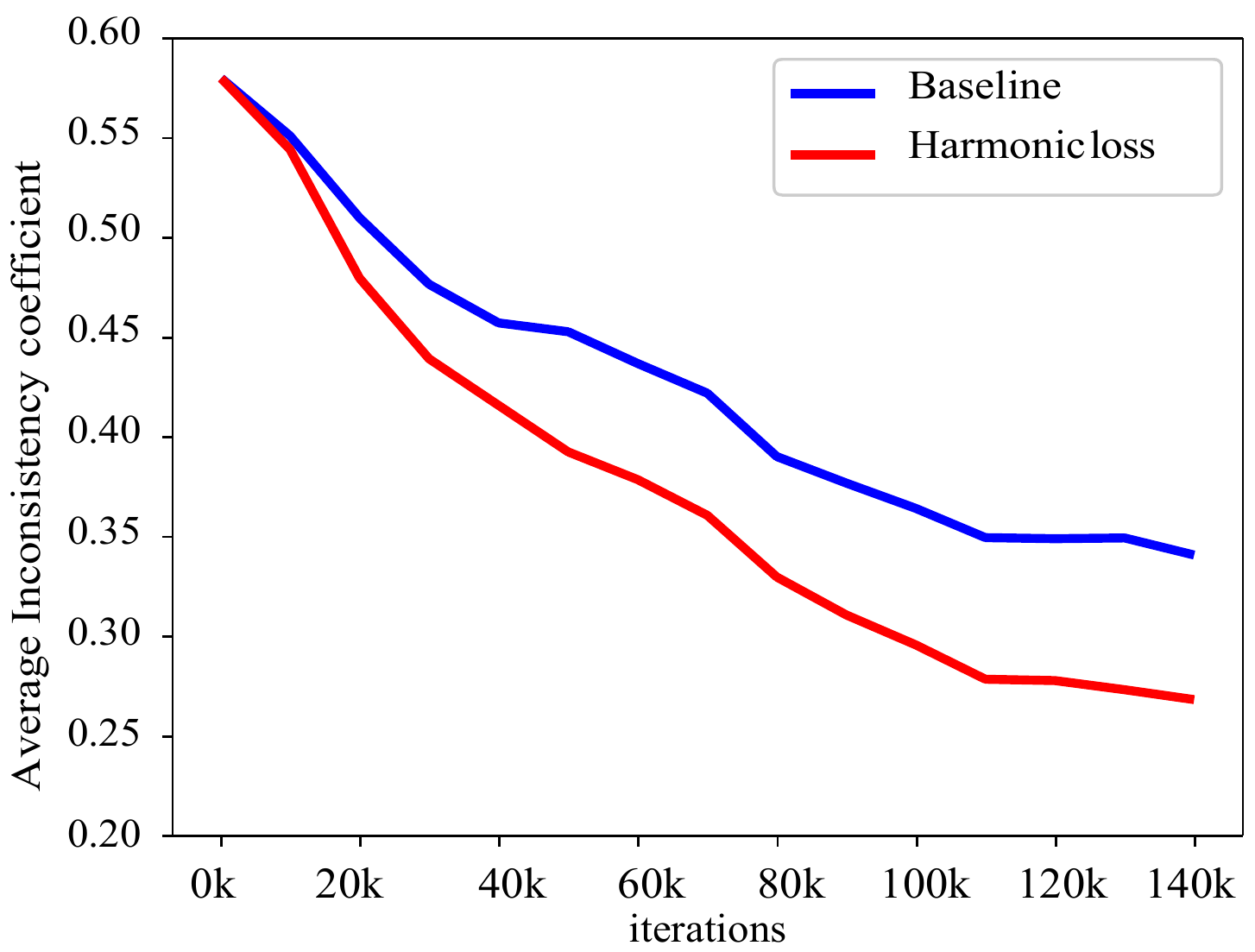}
  \end{minipage}
  }
  \vspace{-0.3cm}
  \caption{(a) is the distribution of the two harmonic factors (i.e., 1+$\beta_r$, 1+$\beta_c$) during the optimization process. (b) is the distribution of the AICs during the optimization process}
  \label{fig_two} 
  \vspace{-0.3cm}
\end{figure}

\vspace{-0.2cm}
It is clear that we reconcile the training of classification and regression by designing two dynamic harmonic factors in the Harmonic loss. However, \emph{are these two branches optimized harmonically in the actual training process}? In the following, we will give the detailed analysis of prediction consistency during the training phase from two aspects: Firstly, as shown in Fig.~\ref{fig_two} (a), we visualize the dynamic distribution of the two harmonic factors (i.e., 1+$\beta_r$, 1+$\beta_c$) during the optimization process. Obviously, in the early stage of training, the gap between the two harmonic factors is relatively large because the inconsistency of the two branches is serious. But with the advancement of optimization, the gap between the two harmonic factors gradually decreases and remains stable at the end of the training, which means that the two branches are progressively in harmony. In addition, in order to quantitatively analyze the prediction inconsistency between classification and regression during the optimization, we propose a simple metric named Average Inconsistency Coefficient (AIC), which is defined as follows:
\vspace{-0.2cm}
\begin{small}
\begin{equation}
\vspace{-0.2cm}
\begin{split}
\label{eq_met}
AIC=\sum\limits_{i \in Pos}^N |P_i-IoU_i|
\end{split}
\vspace{-0.00cm}
\end{equation}
\end{small}
where \emph{N} is the number of \emph{Positive} samples. $p_i$ and $IoU_i$ denotes the predicted classification score and localization accuracy for the sample $x_i$, resp. Clearly, the smaller AIC value means the smaller prediction inconsistency, and vice versa. We visualize the AICs of the detectors trained with standard detection loss and our Harmonic loss in Fig.~\ref{fig_two} (b). We can find that the AIC of the detector trained with Harmonic loss is significantly smaller than the model trained with the standard detection loss during the training. This also clearly means that our Harmonic loss can reconcile prediction inconsistency between classification and regression.
\vspace{-0.1cm}
\subsection{Ablation Study}
\vspace{-0.2cm}
We perform a thorough ablation study on each component of our method. The Faster R-CNN detector is adopted to perform ablation experiments on MS COCO dataset. Table~\ref{sample-table3} shows the effects of each component of our method.
\vspace{-0.1cm}
\begin{table}
  \caption{Ablation results of each component (i.e., Harmonic loss and Harmonic IoU (HIoU) loss) of our methods on MS COCO \texttt{val} set. Faster R-CNN is adopted as the baseline.}
  \footnotesize
  \label{sample-table3}
  \centering
  \begin{tabular}{c|p{0.4cm}p{0.35cm}p{0.35cm}p{0.35cm}p{0.35cm}p{0.3cm}}

    \hline

     \multicolumn{1}{c|}{method }              &  AP& AP$_{50}$ & AP$_{75}$ & AP$_{S}$ & AP$_{M}$ & AP$_{L}$   \\
    \hline
     \multicolumn{1}{c|}{\footnotesize Baseline(Faster R-CNN) }      &37.3 & 58.6  &40.7&21.0&40.1&49.2  \\
     \multicolumn{1}{c|}{\footnotesize +Harmonic loss}   &38.5&59.3&41.8&21.7&41.4&50.1\\
     \multicolumn{1}{c|}{\footnotesize +Harmonic loss+IoU loss } &38.7&59.3&42.1&21.6&41.8&50.3\\
     \multicolumn{1}{c|}{\footnotesize +Harmonic loss+HIoU loss }  & \textbf{39.2}&59.2&42.7&22.0&42.0&50.9\\
    \hline
  \end{tabular}
  \vspace{-0.3cm}
\end{table}

\begin{figure}[t]
 \centering
 \begin{overpic}[width=1\linewidth,height=3cm]{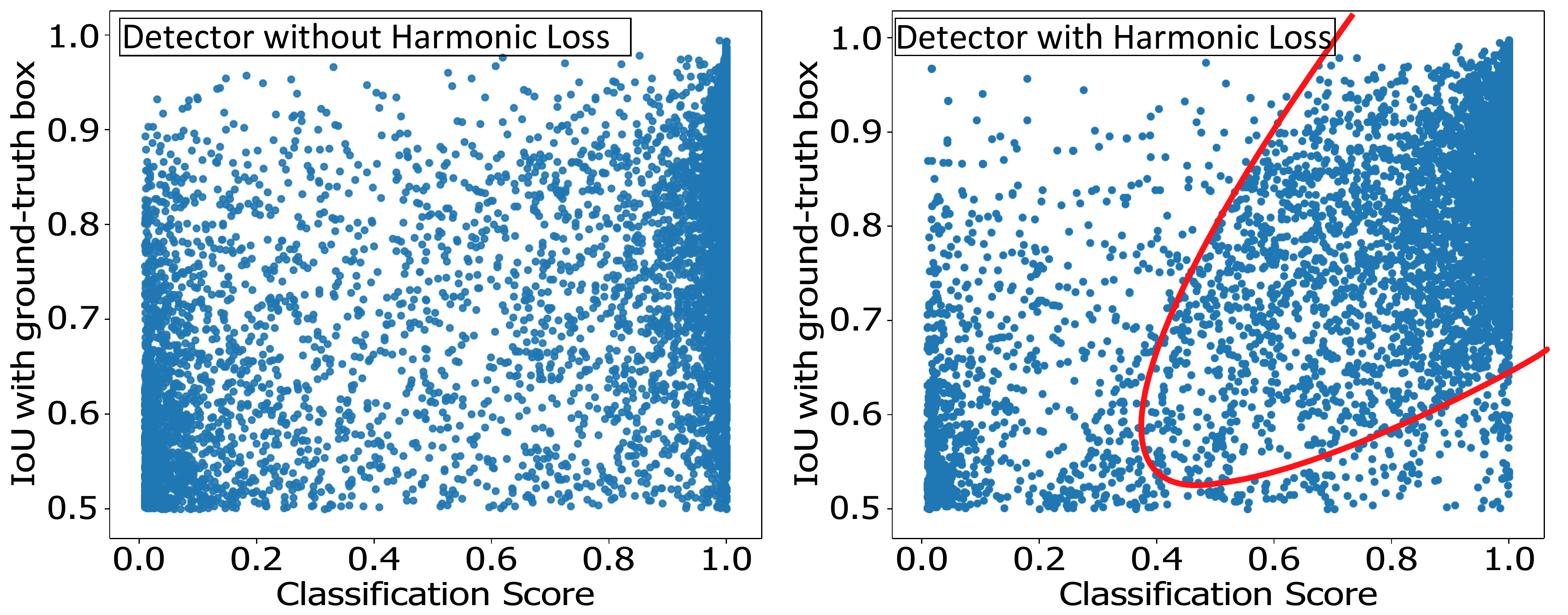}
 \put(25,-3.4){\scriptsize (a)}
 \put(74,-3.4){ \scriptsize(b)}
 \end{overpic}
 \vspace{-0.15cm}
 \caption{The distribution of classification scores and IoUs of all detected bboxes. The red curve is an indicator curve, and the detected bboxes at the upper right corner of the curve are harmonious detection results with high classification score and high IoU.}
 \label{gaijin}
 \vspace{-0.2cm}
\end{figure}

\begin{table}
  \caption{Ablation results of the TC loss in our Harmonic loss}
  \footnotesize
  \label{sample-table4}
  \centering
  \begin{tabular}{c|p{0.5cm}p{0.5cm}p{0.5cm}p{0.5cm}p{0.5cm}p{0.5cm}}

    \hline

     \multicolumn{1}{c|}{Harmonic loss }              &  AP& AP$_{50}$ & AP$_{75}$ & AP$_{S}$ & AP$_{M}$ & AP$_{L}$  \\
    \hline
     \multicolumn{1}{c|}{\footnotesize Without TC Loss}   &38.3&59.0&41.7&21.5&41.1&49.8\\
     \multicolumn{1}{c|}{\footnotesize With TC Loss}   &38.5&59.3&41.8&21.7&41.4&50.1\\
    \hline
  \end{tabular}
  \vspace{-0.5cm}
\end{table}
\textbf{Analysis of Harmonic Loss}. As shown in Table~\ref{sample-table3}, we first conduct the ablation experiments on the MS COCO dataset. Our Harmonic loss can improve the AP by 1.2$\%$ (from 37.3$\%$ to 38.5$\%$). The detection result of each IoU level has been improved, and as the IoU increases, the improvement of AP is more significant. This indicates that the model trained with our Harmonic loss can output harmonious boxes with high co-occurrence of top classification and localization. We further analyze the effect of the TC loss in the Harmonic loss in Table~\ref{sample-table4}. We can find that the APs can be improved by a small margin with low IoU thresholds, which indicates that the TC loss can convert some negative detection results with poor localization into positive detection results with IoU threshold greater than 0.5. Since the harmonic loss is proposed to reconcile classification task and regression task, but \emph{does it really alleviate misalignment between the classification score and the localization accuracy in the testing phase}? In Fig.~\ref{gaijin}, we visualize the distribution of classification scores and IoUs of all detected bounding boxes. Compared with detector without Harmonic loss (a), more predicted bounding boxes of our improved detector (b) appear inside the red curve, which means that our detector can predict more bounding boxes with both high classification scores and high regression accuracy (IoU). In other words, our detector clearly alleviates the disharmony between classification score and localization accuracy.
\begin{table}[t]
  \caption{Ablation results of the $\alpha$ and $\gamma$ in HIoU loss.}
  \footnotesize
  \label{sample-table5}
  \centering
  \begin{tabular}{cc|p{0.7cm}p{0.6cm}p{0.6cm}p{0.6cm}p{0.6cm}p{0.6cm}}

    \hline

     \small$\alpha$&    \small$\gamma$          &AP& AP$_{50}$ & AP$_{75}$ & AP$_{S}$ & AP$_{M}$ & AP$_{L}$  \\
    \hline
     0.5& 0.8  &38.9&59.3&42.1&21.9&41.7&50.5\\
     1.0& 0.8 &39.0&59.3&42.3&22.1&41.7&50.8\\
     1.5& 0.8  &\textbf{39.2}&59.2&42.7&22.0&42.0&50.9\\
     2.5& 0.8  &38.9&58.6&43.0&21.6&41.8&50.6\\
     \hline
     1.5& 0.0  &38.7 &59.3&42.1 &21.6&41.8&50.3\\
     1.5& 0.5  &38.9&59.2&42.4&21.7&41.8&50.7\\
     1.5& 0.8  &\textbf{39.2}&59.2&42.7&22.0&42.0&50.9\\
     1.5& 1.0  &39.0&58.7&42.8&21.9&41.9&50.9\\
    \hline
  \end{tabular}
  \vspace{-0.5cm}
\end{table}

\textbf{Analysis of HIoU Loss}. In order to validate the effectiveness of our HIoU loss, we conduct three ablation experiments. \emph{Firstly}, we add the standard IoU loss and HIoU loss to the Harmonic loss respectively. As shown in Table~\ref{sample-table3}, when we add the standard IoU loss into Harmonic loss, the AP can only be improved by a small margin (from 38.5$\%$ to 38.7$\%$). But when we add the HIoU loss into Harmonic loss, our method can achieve 0.7$\%$ improvement (from 38.5$\%$ to 39.2$\%$) compared to the detector without HIoU loss. Specifically, the improvement for AP at higher IoU threshold (e.g., 0.75) is very significant, which means that our HIOU loss prevents the localization loss from being dominated by outliers in training phase. The reason is that our HIoU loss can alleviate the imbalance bias of samples with different IoU levels, as shown in Fig.\ref{zhu} (b). \emph{Secondly}, we conduct experiments to analyze the effects of the focusing parameter $\gamma$ of the HIoU loss. As shown in Table~\ref{sample-table5}, we can get the best AP when we set the focusing parameter $\gamma$ to 0.8.
\emph{Thirdly}, we also conduct experiments to analyze the effects of the trade-off $\alpha$ between SmoothL1 loss and HIoU loss. As shown in Table~\ref{sample-table5}, we experiment with four different ratios ranging from 0.5 to 2.5 and get the best result when $\alpha=1.5$. So we adopt $\alpha=1.5$, $\gamma=0.8$ in our experiments.
\vspace{-0.1cm}
\subsection{Qualitative Detection of Diverse Objects}
\begin{figure}[t]
\begin{center}
\vspace{-0.2cm}
\includegraphics[width=0.60\linewidth]{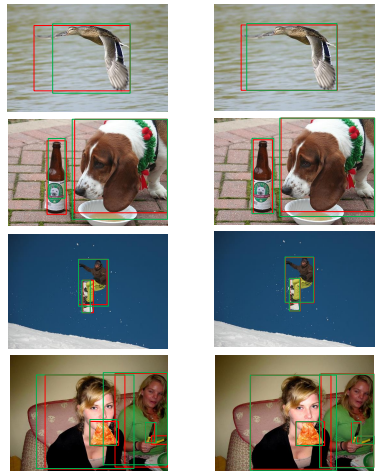}
\end{center}
\vspace{-0.4cm}
\caption{Qualitative comparisons between the standard detection loss and our HarmonicDet approach based on the Faster R-CNN. Several examples of irregular and occluded object detection are illustrated. The first column shows the results of Faster R-CNN trained with standard detection loss. The second column shows the results of Faster R-CNN trained with HarmonicDet loss. Red boxes represent the GT and green boxes represent the predicted boxes.}
\vspace{-8pt}
\label{keshihua}
\end{figure}

\vspace{-0.2cm}
Qualitative comparisons between the standard Faster R-CNN and our method are provided in Fig.~\ref{keshihua}. We can clearly find that, compared to Faster R-CNN trained with standard detection loss, the Faster R-CNN trained with our HarmonicDet loss can obtain more accurate detected bounding boxes. This indicates that by using HarmonicDet loss in the training phase to strengthen the correlation between the two tasks, the disharmony between classification score and localization accuracy can be effectively alleviated in the testing phase.
\vspace{-0.2cm}
\section{Conclusion}
\vspace{-0.3cm}
In this paper, in order to reconcile prediction consistency between classification and regression, we propose a Harmonic loss to harmonize the training of classification and regression in object detection. The Harmonic loss enables classification task and the localization task to dynamically promote each other in the training phase, thereby producing the consistent bounding boxes with high co-occurrence of top classification and localization in the testing phase. Furthermore, in order to prevent the localization loss from being dominated by outliers during training phase, a Harmonic IoU (HIoU) loss is proposed to harmonize the weight of the localization loss of different IoU-level samples. Comprehensive experiments by plugging and playing the proposed loss into six popular detectors demonstrate the generality and effectiveness of our HarmonicDet approach. 

\textbf{Acknowledgement.} This work was supported by the National Science Fund of China under Grants (61771079) and Fundamental Scientific Research Fund of Central Universities (No. 2020CDCGTX061).

{\small
\bibliographystyle{ieee_fullname}
\bibliography{egbib}
}

\end{document}